\documentclass[conference,a4paper]{IEEEtran}

\ifCLASSINFOpdf
\else
\fi
\usepackage{amsmath,amssymb,amsthm, bm}
\usepackage[dvipdfmx]{graphicx}
\usepackage{enumerate}
\usepackage{my-style}

\usepackage[top=19.1truemm,bottom=30.1truemm,left=13.1truemm,right=13.1truemm]{geometry}

\begin{document}
%
\title{Effective Mean-Field Inference Method for Nonnegative Boltzmann Machines}

\author{\IEEEauthorblockN{Muneki Yasuda}
\IEEEauthorblockA{Graduate School of Science and Engineering, Yamagata University\\
Jyounan 4-3-16, Yonezawa, Yamgata pref., 992-8510, Japan\\
Email: muneki@yz.yamagata-u.ac.jp}
}

%


\maketitle

\begin{abstract}
Nonnegative Boltzmann machines (NNBMs) are recurrent probabilistic neural network models  
that can describe multi-modal nonnegative data. 
NNBMs form rectified Gaussian distributions that appear in biological neural network
models, positive matrix factorization, nonnegative matrix factorization, and so on. 
In this paper, an effective inference method for NNBMs is proposed that uses the mean-field method, referred to as the Thouless--Anderson--Palmer equation, 
and the diagonal consistency method, which was recently proposed.
\end{abstract}


%
\IEEEpeerreviewmaketitle

\section{Introduction} \label{sec:introduction}

Techniques for inferencing statistical values on Markov random fields (MRFs) are fundamental techniques used in various fields 
involving pattern recognition, machine learning, and so on. 
The computation of statistical values on MRFs is generally intractable because of exponentially increasing computational costs.
\textit{Mean-field methods}, which originated in the field of statistical physics, are the major techniques for computing statistical values approximately. 
Loopy belief propagation methods are the most widely-used techniques in the mean-field methods~\cite{Pearl1988,Yedidia&Freeman2005}, 
and various studies on them have been conducted.  

Many mean-field methods on MRFs with discrete random variables have been developed thus far. 
However, as compared with the discrete cases,
mean-field methods on MRFs with continuous random variables have not been developed well. 
One reason for this is thought to be that the application of loopy belief propagation methods to continuous MRFs is not straightforward 
and has not met with no success except in Gaussian graphical models~\cite{Yedidia&Freeman2001}. 
Although it is possible to approximately construct a loopy belief propagation method on a continuous MRF by approximating 
continuous functions using histograms divided by bins, this is not practical.

\textit{Nonnegative Boltzmann machines} (NNBMs) are recurrent probabilistic neural network models  
that can describe multi-modal nonnegative data~\cite{Downs&Mackay&Lee2000}. 
For continuous nonnegative data, an NNBM is expressed in terms of a maximum
entropy distribution that matches the first and second order statistics of the data. 
Because random variables in NNBMs are bounded, 
NNBMs are not standard multivariate Gaussian distributions, but 
\textit{rectified Gaussian distributions} that appear in biological neural network
models of the visual cortex, motor cortex, and head direction system~\cite{Socci&Lee&Seung1998},  
positive matrix factorization~\cite{Paatero&Tapper1994}, nonnegative
matrix factorization~\cite{Lee&Seung1999}, nonnegative independent
component analysis~\cite{Plumbley&Oja2004}, and so on. 
A learning algorithm for NNBMs was formulated by means of a variational Bayes method~\cite{Harva&Kaban2007}. 

Because NNBMs are generally intractable, 
we require an effective approximate approach in order to provide probabilistic inference and statistical learning algorithms for the NNBM.
A mean-field method for NNBMs was first proposed in reference \cite{Downs&Mackay&Lee2000}. 
This method corresponds to a naive mean-field approximation, which is the most basic mean-field technique. 
A higher-order approximation than the naive mean-field approximation for NNBMs was proposed by Downs~\cite{Downs2001}
and recently, his method was improved by the author~\cite{Yasuda&Tanaka2012}.
These methods correspond to \textit{the Thouless--Anderson--Palmer} (TAP) equation in statistical physics, 
and they are constructed using a perturbative approximative method referred to in statistical physics as the Plefka expansion~\cite{Plefka1982}. 
Recently, a mean-field method referred to as \textit{the diagonal consistency method} was proposed~\cite{I-SusP2013}. 
The diagonal consistency method is a powerful method that can increase the performance of mean-field methods by using a simple extension for them~\cite{I-SusP2013,Raymond2013,Yasuda2013}. 

In this paper, we propose an effective mean-field inference method for NNBMs in which the TAP equation, proposed in reference \cite{Yasuda&Tanaka2012}, 
is combined with the diagonal consistency method. 
The remainder of this paper is organized as follows. 
In section \ref{sec:NNBM} the NNBM is introduced, and subsequently, the TAP equation for the NNBM is formulated in section \ref{sec:Gibbs&TAP} 
in accordance with reference \cite{Yasuda&Tanaka2012}.   
The proposed methods are described in section \ref{sec:propose}. 
The main proposed method combined with the diagonal consistency method is formulated in section \ref{subsec:I-SusP}. 
In section \ref{sec:numerical}, we present some numerical results of statistical machine learning on NNBMs and verify the performance of the proposed method numerically. 
Finally, in section \ref{sec:conclusion} we present our conclusions.

\section{Nonnegative Boltzmann Machines}\label{sec:NNBM}

In probabilistic information processing, various applications involve continuous nonnegative data, e.g., digital image filtering, automatic human face recognition, etc.
In general, standard Gaussian distributions, which are unimodal distributions, are not good approximations for such data.
Nonnegative Boltzmann machines are probabilistic machine learning models that can describe multi-modal nonnegative data,  
and were developed for the specific purpose of modeling continuous nonnegative data~\cite{Downs&Mackay&Lee2000}.

Let us consider an undirected graph $G(V,E)$, where $V=\{i \mid i = 1,2,\ldots,n\}$ is the set of vertices in the graph and $E=\{(i,j)\}$ is the set of undirected links. 
On the undirected graph, an NNBM is defined by the Gibbs-Boltzmann distribution,
\begin{align}
P(\bm{x}\mid \bm{b},\bm{w}):=\frac{1}{Z(\bm{b},\bm{w})} \exp \big(-E(\bm{x};\bm{b},\bm{w})\big),
\label{eq:NNBM}
\end{align}
where 
\begin{align*}
E(\bm{x};\bm{b},\bm{w}):=-\sum_{i=1}^n b_i x_i + \frac{1}{2}\sum_{i=1}^n w_{ii}x_i^2 +\sum_{(i,j) \in E} w_{ij} x_ix_j
\end{align*}
is the energy function. The second summation in the energy function is taken over all the links in the graph. 
The expression $Z(\bm{b},\bm{w})$ is the normalized constant called the partition function.
The notations $\bm{b}$ denote the bias parameters and the notations $\bm{w}$ denote the coupling parameters of the NNBM.
All variables $\bm{x}$ take nonnegative real values, and $w_{ii}>0$ and $w_{ij}=w_{ji}$ are assumed. 
The distribution in equation (\ref{eq:NNBM}) is called the rectified Gaussian distribution, and 
it can represent a multi-modal distribution, unlike standard Gaussian distributions \cite{Socci&Lee&Seung1998}. 
In the rectified Gaussian distribution, the symmetric matrix $\bm{w}$ is \textit{a co-positive matrix}.
The set of co-positive matrices is larger than the set of positive definite matrices that can be used in a standard Gaussian distribution, 
and is employed in co-positive programming.

\section{variational free energy and the TAP equation for the NNBM} \label{sec:Gibbs&TAP}

To lead to the TAP equation for the NNBM, which is an alternative form of the TAP equation proposed in reference \cite{Yasuda&Tanaka2012}, 
and a subsequent proposed approximate method for the NNBM in this paper, 
let us introduce a variational free energy, which is called the Gibbs free energy in statistical physics, for the NNBM. 
The variational free energy of the NNBM in equation (\ref{eq:NNBM}) is expressed by
\begin{align}
&\mcal{F}(\bm{m},\bm{v};\bm{b},\bm{w}):=-\sum_{i=1}^n b_i m_i+\frac{1}{2}\sum_{i=1}^n w_{ii}v_i\nn
&+\max_{\bm{l},\bm{r}}\Big\{ \sum_{i=1}^n l_i m_i -\frac{1}{2}\sum_{i=1}^n r_iv_i-\ln \mcal{Z}(\bm{l},\bm{r},\bm{w}) \Big\},
\label{eq:GibbsFreeEnergy_NNBM}
\end{align}
where
\begin{align*}
&\mcal{Z}(\bm{l},\bm{r},\bm{w})\nn
&:=\int_0^{\infty}\exp\Big(\sum_{i=1}^n l_i x_i - \frac{1}{2}\sum_{i=1}^n r_{i}x_i^2 -\sum_{(i,j) \in E} w_{ij} x_ix_j\Big)\diff\bm{x}.
\end{align*}
A brief derivation of this variational free energy is given in Appendix \ref{app:derivation-VFE}.
The notations $\bm{l}=\{l_i \mid i \in V\}$ and $\bm{r}=\{r_i \mid i \in V\}$ are variables that originated in the Lagrange multiplier.
It should be noted that the independent variables $m_i$ and $v_i$ represent the expectation value and the variance of $x_i$, respectively. 
In fact, the values of $\bm{m}=\{m_i \mid i \in V\}$ and $\bm{v}=\{v_i \mid i \in V\}$, 
which minimize the variational free energy, coincide with expectations, $\ave{x_i}$, and variances, $\ave{x_i^2}$, on the NNBM, respectively, 
where $\ave{\cdots}:= \int_0^{\infty}(\cdots)P(\bm{x}\mid \bm{b},\bm{w})\diff\bm{x}$. 
The minimum of the variational free energy is equivalent to $-\ln Z(\bm{b},\bm{w})$.

However, the minimization of the variational free energy is intractable, because it is not easy to evaluate the partition function $\mcal{Z}(\bm{l},\bm{r},\bm{w})$.
Therefore, an approximative instead of the exact minimization. 
By using the Plefka expansion~\cite{Plefka1982}, the variational free energy can be expanded as
\begin{align}
\mcal{F}(\bm{m},\bm{v};\bm{b},\bm{w})=\mcal{F}_{\mrm{2nd}}(\bm{m},\bm{v};\bm{b},\bm{w}) + O(w_{ij}^3),
\label{eq:PlefkaExpansion}
\end{align}
where
\begin{align}
&\mcal{F}_{\mrm{2nd}}(\bm{m},\bm{v};\bm{b},\bm{w})\nn
&=-\sum_{i = 1}^n b_i m_i+\frac{1}{2}\sum_{i=1}^n w_{ii}v_i
 + \sum_{i=1}^n\Big\{l_i^* m_i - \frac{1}{2} r_i^*v_i\nn
& - \ln \mrm{erfcx}\Big(-\frac{l_i^*}{\sqrt{2r_i^*}}\Big)- \frac{1}{2}\ln\frac{\pi}{2 r_i^*}\Big\} \nn
&+\sum_{(i,j)\in E}w_{ij}m_im_j -\frac{1}{2}\sum_{(i,j)\in E}w_{ij}^2 \big(v_{i}-m_i^2\big)\big( v_{j}-m_j^2\big),
\label{eq:Gibbs_NNBM_2nd}
\end{align}
where the function $\mrm{erfcx}(x)$ is the scaled complementary error function defined by
\begin{align*}
\mrm{erfcx}(x):= \frac{2e^{x^2}}{\sqrt{\pi}} \int_x^{\infty} e^{-z^2}\diff z.
\end{align*}
Equation (\ref{eq:PlefkaExpansion}) is the perturbative expansion of the variational free energy with respect to $w_{ij}\> (i\not=j)$.
The notations $l_i^*$ and $r_i^*$ are defined by 
\begin{align}
\{l_i^*,r_i^*\}
&=\argmax_{l_i,r_i}\Big\{l_i m_i - \frac{1}{2} r_i v_i - \ln \mrm{erfcx}\Big(-\frac{l_i}{\sqrt{2 r_i}}\Big)\nn
\aleq + \frac{1}{2}\ln r_i\Big\}.
\label{eq:def-l&r_2nd}
\end{align}
Let us neglect the higher-order terms and approximate the true variational free energy, $\mcal{F}(\bm{m},\bm{v};\bm{b},\bm{w})$, 
by the variational free energy in equation (\ref{eq:Gibbs_NNBM_2nd}).
The minimization of the approximate variational free energy in equation (\ref{eq:Gibbs_NNBM_2nd}) is straightforward. 
It can be done by computationally solving the nonlinear simultaneous equations,
\begin{align}
m_i &= \frac{l_i^*}{r_i^*} + \sqrt{\frac{2}{\pi r_i^*}}\mrm{erfcx}\Big(-\frac{l_i^*}{\sqrt{2 r_i^*}}\Big)^{-1},
\label{eq:mi_NNBM_2nd}
\\
v_i&=\big(1 + l_i^*m_i\big) / r_i^*,
\label{eq:vi_NNBM_2nd}
\\
l_i^*&= b_i- \sum_{j \in \partial(i)}w_{ij}m_j- \sum_{j \in \partial(i)}w_{ij}^2m_i\big( v_{j}-m_j^2\big),
\label{eq:l_NNBM_2nd}
\\
r_i^*&=w_{ii} - \sum_{j \in \partial(i)}w_{ij}^2 \big( v_{j}-m_j^2\big),
\label{eq:r_NNBM_2nd}
\end{align}
by using an iteration method, such as a successive iteration method.
The notation $\partial(i)$ is the set of nearest-neighbor vertices of vertex $i$. 
Equations (\ref{eq:l_NNBM_2nd}) and (\ref{eq:r_NNBM_2nd}) come from the minimum condition of equation (\ref{eq:Gibbs_NNBM_2nd}) with respect to $\bm{m}$ and $\bm{v}$, respectively, 
and equations (\ref{eq:mi_NNBM_2nd}) and (\ref{eq:vi_NNBM_2nd}) come from the maximum conditions in equation (\ref{eq:def-l&r_2nd}). 

Equations (\ref{eq:mi_NNBM_2nd})--(\ref{eq:r_NNBM_2nd}) are referred to as the TAP equation for the NNBM. 
Although the expression of this TAP equation is different from the expression proposed in reference \cite{Yasuda&Tanaka2012}, 
they are essentially almost the same.
Since we neglect higher-order effects in the true variational free energy, 
the expectation values and the variances obtained by the TAP equation are approximations, except in special cases. 
It is known that, 
on NNBMs which are defined on complete graphs whose sizes are infinitely large 
and whose coupling parameters $\bm{w}$ are distributed 
in accordance with a Gaussian distribution whose variance is scaled by $n^{-1}$, 
the TAP equation gives true expectations~\cite{Yasuda&Tanaka2012}.

\section{Proposed Method} \label{sec:propose}

In this section, we propose an effective mean-field inference method for the NNBM in which the TAP equation in equations (\ref{eq:mi_NNBM_2nd})--(\ref{eq:r_NNBM_2nd}) 
is combined with the the diagonal consistency method proposed in reference \cite{I-SusP2013}. 
The diagonal consistency method is a powerful method that can increase the performance of mean-field methods by using a simple extension for them. 

\subsection{Liner response relation and susceptibility propagation method} \label{subsec:LRR&SusP}

Before addressing the proposed method, let us formulate \textit{the linear response relation} for the NNBM. 
The linear response relation enables us to evaluate higher-order statistical values, such as covariances. 
Here, we define the matrix $\bm{\chi}$ as
\begin{align}
M_{ij}:= \frac{\partial m_i^*}{\partial b_j}, 
\label{eq:LRR}
\end{align}
where $\bm{m}^*=\{m_i^*\mid i \in V\}$ is the values of $\bm{m}$ that minimize the (approximate) variational free energy. 
Since relations $M_{ij}=\ave{x_ix_j}-\ave{x_i}\ave{x_j}$ exactly hold when $\bm{m}^*$ are the true expectation values,  
we can interpret the matrix $\bm{M}$ as the covariant matrix on the NNBM. 
The quantity $M_{ij}$ is referred to as \textit{the susceptibility} in physics, 
and is interpreted as a response on vertex $i$ for a quite small change of the bias on vertex $j$. 
In practice, we use $\bm{m}^*$ obtained by a mean-field method instead of the intractable true expectations 
and approximately evaluate $\bm{M}$ by means of the mean-field method.

Let us use $\bm{m}^*$ obtained by the minimization of the approximate variational free energy in equation (\ref{eq:Gibbs_NNBM_2nd}), 
that is the solutions to the TAP equation in equations (\ref{eq:mi_NNBM_2nd})--(\ref{eq:r_NNBM_2nd}), in equation (\ref{eq:LRR}) 
in order to approximately compute the susceptibilities. 
By using equations (\ref{eq:mi_NNBM_2nd})--(\ref{eq:LRR}), 
the approximate susceptibilities are obtained by using the solutions to the nonlinear simultaneous equations, which are
\begin{align}
M_{ij}&=\big(v_i-m_i^2\big) L_{ij} - \frac{1}{2r_i^*}\big( l_i^* v_i - m_i^2l_i^* + m_i\big)R_{ij},
\label{eq:SusP-chi}\\
V_{ij}&=\frac{ m_i L_{ij} + l_i^*M_{ij}-v_i R_{ij} }{r_i^*},
\label{eq:SusP-V}\\
L_{ij}&=\delta_{i,j}-\sum_{k \in \partial(i)}w_{ik}M_{kj} \nn
\aleq
- \sum_{k \in \partial(i)}w_{ik}^2 \big\{ M_{ij}\big( v_{k}-m_k^2\big) + m_i\big( V_{kj} - 2m_k M_{kj}\big)\big\},
\label{eq:SusP-L}\\
R_{ij}&=-\sum_{k \in \partial(i)}w_{ik}^2\big( V_{kj} - 2m_k M_{kj}\big),
\label{eq:SusP-R}
\end{align}
where $V_{ij}:= \partial v_i/ \partial b_j$, $L_{ij}:= \partial l_i^*/ \partial b_j$, 
and $R_{ij}:=\partial r_i^* / \partial b_j$. The expression $\delta_{i,j}$ is the Kronecker delta.
The variables $\bm{m}$, $\bm{v}$, $\bm{l}^*$, and $\bm{r}^*$ in equations (\ref{eq:SusP-chi}) and (\ref{eq:SusP-V}) are the solutions to 
the TAP equation in equations (\ref{eq:mi_NNBM_2nd})--(\ref{eq:r_NNBM_2nd}).
Equations (\ref{eq:SusP-chi})--(\ref{eq:SusP-R}) are obtained by differentiating equations (\ref{eq:mi_NNBM_2nd})--(\ref{eq:r_NNBM_2nd}) with respect to $b_j$. 
Note that, the expression in equations (\ref{eq:SusP-chi}) and (\ref{eq:SusP-V}) is simplified by using equations (\ref{eq:mi_NNBM_2nd}) and (\ref{eq:vi_NNBM_2nd}). 

After obtaining solutions to the TAP equation in equations (\ref{eq:mi_NNBM_2nd})--(\ref{eq:r_NNBM_2nd}), 
by computationally solving equations (\ref{eq:SusP-chi})--(\ref{eq:SusP-R}) by using an iterative method, such as the successive iteration method, 
we can obtain approximate susceptibilities $\bm{M}$ in terms of the TAP equation. 
This scheme for computing susceptibilities has a decade-long history and is referred to as \textit{the susceptibility propagation method}
or to as \textit{the variational liner response method}~\cite{KTanaka2003, Welling&Teh2003, Welling&Teh2004, Mezard&Mora2009}. 
By using this method, we can obtain not only the expectation values of one variable but also the covariances on NNBMs.
In the next section, we propose a version of this susceptibility propagation method that is improved by combining it with the diagonal consistency method.

\subsection{Improved susceptibility propagation method for the NNBM} \label{subsec:I-SusP}

In order to apply the diagonal consistency method to the present framework, I modify the approximate variational free energy 
in equation (\ref{eq:Gibbs_NNBM_2nd}) as
\begin{align}
\mcal{F}_{\mrm{2nd}}^{\dagger}(\bm{m},\bm{v};\bm{b},\bm{w}, \bm{\Lambda})&:=\mcal{F}_{\mrm{2nd}}(\bm{m},\bm{v};\bm{b},\bm{w}) \nn
\aldef\>-\frac{1}{2}\sum_{i=1}^n\Lambda_i \big(v_i-m_i^2\big),
\label{eq:new-Gibbs_NNBM_2nd}
\end{align}
where the variables $\bm{\Lambda}=\{\Lambda_i \mid i \in V\}$ are auxiliary independent parameters 
that play an important role in the diagonal consistency method~\cite{I-SusP2013, Yasuda2013}. 
We formulate the TAP equation and the susceptibility propagation for the modified variational free energy, 
$\mcal{F}_{\mrm{2nd}}^{\dagger}(\bm{m},\bm{v};\bm{b},\bm{w}, \bm{\Lambda})$, in the same manner as that described in the previous sections.
By this modification, equations (\ref{eq:l_NNBM_2nd}), (\ref{eq:r_NNBM_2nd}), and (\ref{eq:SusP-L}) are changed to 
\begin{align}
l_i^*&= b_i - \sum_{j \in \partial(i)}w_{ij}m_j- \sum_{j \in \partial(i)}w_{ij}m_i\big( v_{j}-m_j^2\big)\nn
\aleq -m_i\Lambda_i,
\label{eq:l_NNBM_2nd-new}\\
r_i^*&= w_{ii} - \sum_{j \in \partial(i)}w_{ij}^2 v_j-\Lambda_i,
\label{eq:r_NNBM_2nd-new}
\end{align}
and
\begin{align}
&L_{ij}=\delta_{ij}-\sum_{k \in \partial(i)}w_{ik}M_{kj} \nn
&- \sum_{k \in \partial(i)}w_{ik}^2 \big\{ M_{ij}\big( v_{k}-m_k^2\big) + m_i\big( V_{kj} - 2m_k M_{kj}\big)\big\}\nn
&-M_{ij}\Lambda_i,
\label{eq:SusP-L-new}
\end{align}
respectively. Equations (\ref{eq:mi_NNBM_2nd})--(\ref{eq:l_NNBM_2nd}) and (\ref{eq:r_NNBM_2nd-new}) are the TAP equation, 
and equations (\ref{eq:SusP-chi}), (\ref{eq:SusP-V}), (\ref{eq:SusP-R}), and (\ref{eq:SusP-L-new}) are the susceptibility propagation for the modified variational free energy. 
The solutions to these equations obviously depend on the values of $\bm{\Lambda}$. 
Since the values of $\bm{\Lambda}$ cannot be specified by using only the TAP equation and the susceptibility propagation, 
we need an additive constraint in order to specify them.

As mentioned in sections \ref{sec:Gibbs&TAP} and \ref{subsec:LRR&SusP}, 
relations $v_i = \ave{x_i^2}$ and $M_{ii}=\ave{x_i^2}-\ave{x_i}^2$ hold in the scheme with no approximation.
Hence, the relations 
\begin{align}
M_{ii} = v_i - m_i^2
\label{eq:diagonal-consistency}
\end{align}
should always hold in the exact scheme, and therefore, these relations can be regarded as important information that the present system has. 
The diagonal consistency method is a technique for inserting these diagonal consistency relations into the present approximation scheme. 
In the diagonal consistency method, the values of the auxiliary parameters $\bm{\Lambda}$ are determined 
so that the solutions to the modified TAP equation in equations (\ref{eq:mi_NNBM_2nd})--(\ref{eq:l_NNBM_2nd}) and (\ref{eq:r_NNBM_2nd-new}) 
and the modified susceptibility propagation in equations (\ref{eq:SusP-chi}), (\ref{eq:SusP-V}), (\ref{eq:SusP-R}), and (\ref{eq:SusP-L-new})
satisfy the diagonal consistency relations in equation (\ref{eq:diagonal-consistency}).
From equations (\ref{eq:SusP-chi}), (\ref{eq:SusP-L-new}), and (\ref{eq:diagonal-consistency}), we obtain
\begin{align}
\Lambda_i=\frac{A_i}{v_i -m_i^2} + \frac{B_i}{(v_i-m_i^2)^2},
\label{eq:diagMatch}
\end{align} 
where
\begin{align*}
A_i&:=-\sum_{k \in \partial(i)}w_{ik}\chi_{ki} \nn
&- \sum_{k \in \partial(i)}w_{ik}^2 \big\{ \big(v_i -m_i^2\big)\big( v_{k}-m_k^2\big) + m_i\big( V_{ki} - 2m_k \chi_{ki}\big)\big\},\nn
B_i&:=- \frac{1}{2r_i^*}\big( l_i^* v_i - m_i^2l_i^* + m_i\big)R_{ij}.
\end{align*}
Equation (\ref{eq:diagMatch}) is referred to as \textit{the diagonal matching equation}, 
and we determine the values of $\bm{\Lambda}$ by using this equation in the framework of the diagonal consistency method~\cite{I-SusP2013, Yasuda2013}.
By computationally solving the modified TAP equation, the modified susceptibility propagation, and the diagonal matching equation, 
the expectations $\bm{m}$ and susceptibilities (covariances) $\bm{M}$ that we can obtain are improved by applying the diagonal consistency method. 
It is noteworthy that the order of the computational cost of the proposed method is the same as that of the normal susceptibility propagation method, which is $O(n |E|)$.

\begin{figure}[ht]
\begin{center}
\includegraphics[height=4.0cm]{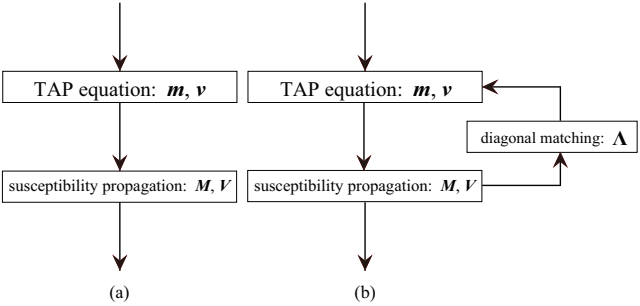}
\end{center}
\caption{Illustration of schemes of (a) normal susceptibility propagation and (b) proposed method. 
The procedure of the normal susceptibility propagation method is straightforward. 
In contrast, the results of the susceptibility propagation method are fed back to the TAP equation through parameters $\bm{\Lambda}$ 
in the improved susceptibility propagation.}
\label{fig:Scheme}
\end{figure}

In the normal scheme presented in section \ref{subsec:LRR&SusP}, 
the results obtained from the susceptibility propagation method are not fed back to the TAP equation. 
In contrast, in the improved scheme proposed in this section, 
they are fed back to the TAP equation through parameters $\bm{\Lambda}$, 
because the TAP equation and the susceptibility propagation method share the parameters $\bm{\Lambda}$ in this scheme (see figure \ref{fig:Scheme}).
Instinctually, the role of the parameters $\bm{\Lambda}$ can be interpreted as follows. 
As mentioned in section \ref{subsec:LRR&SusP}, the susceptibilities $\bm{M}$ are interpreted as responses 
to the small changes in the biases on the vertices. 
Self-response $M_{ii}$, which is a response on vertex $i$ to a small change in the bias on vertex $i$, 
is interpreted as a response coming back through the whole system. 
Thus, it can be expected that the self-responses have some information of the whole system.  
In the proposed scheme, the self-responses, which differ from the true response by the approximation, 
are corrected by the diagonal consistency condition in equation (\ref{eq:diagonal-consistency}), 
and subsequently the information of the whole system, which the self-responses have, are embedded into parameters $\bm{\Lambda}$ and  
are transmitted to the TAP equation through parameters $\bm{\Lambda}$.

\section{Numerical Verification of Proposed Method} \label{sec:numerical}

To verify performance of the proposed method, let us consider statistical machine learning on NNBMs.
For a given complete data set $\mcal{D}=\{\ve{x}^{(\mu)} \mid \mu = 1,2,\ldots, N\}$ generated from a generative NNBM, 
the log-likelihood function of the learning NNBM is defined as
\begin{align*}
\mcal{L}(\bm{b},\bm{w}):=\frac{1}{N}\sum_{\mu=1}^N \ln P(\ve{x}^{(\mu)}\mid \bm{b},\bm{w}).
\end{align*} 
This log-likelihood is rewritten as
\begin{align}
\mcal{L}(\bm{b},\bm{w})&=\sum_{i=1}^n b_i \ave{x_i}_{\mcal{D}} - \frac{1}{2}\sum_{i=1}^n w_{ii}\ave{x_i^2}_{\mcal{D}} \nn
\aleq-\sum_{(i,j) \in E} w_{ij} \ave{x_ix_j}_{\mcal{D}}
+\min_{\bm{m},\bm{v}}\mcal{F}(\bm{m},\bm{v};\bm{b},\bm{w}),
\label{eq:log-likelihood-1}
\end{align}
where notation $\ave{\cdots}_{\mcal{D}}$ denotes the sample average with respect to the given data set.
The learning is achieved by maximizing the log-likelihood function with respect to $\bm{b}$ and $\bm{w}$.
The gradients of the log-likelihood function are expressed as follows:
\begin{align*}
\Delta{b_i}&\propto \ave{x_i}_{\mcal{D}}-m_i,\\
\Delta{w_{ii}}&\propto -\ave{x_i^2}_{\mcal{D}}+v_i,\\
\Delta{w_{ij}}&\propto -\ave{x_ix_j}_{\mcal{D}}+\big( M_{ij} + m_im_j\big).
\end{align*}
We use $\bm{m}$, $\bm{v}$, and $\bm{M}$ obtained by the approximative methods presented in the previous sections in these gradients,  
and approximately maximize the log-likelihood function.  
More accurate estimates can be expected to give better learning solutions. 
We measured the quality of the learning by the mean absolute errors (MAEs) between the true parameters and learning solutions defined by
\begin{align}
e_{\bm{b}}&:=\frac{1}{n}\sum_{i=1}^n \big| b_i^{\mrm{true}} - b_i^{\mrm{learn}} \big|,\nn
e_{\bm{w}}&:=\frac{1}{n+|E|}\Big( \sum_{(i,j)\in E} \big| w_{ij}^{\mrm{true}} - w_{ij}^{\mrm{learn}} \big|\nn
\aldef\>+\sum_{i =1}^n \big| w_{ii}^{\mrm{true}} - w_{ii}^{\mrm{learn}} \big|\Big),
\label{eq:MAEs}
\end{align}
where $\bm{b}^{\mrm{true}}$ and $\bm{w}^{\mrm{true}}$ are the true parameters on the generative NNBM. 
In the following sections, the results of the learning on two kinds of generative NNBMs are shown. 
In both cases, we set to $n=36$ and $N=10000$ and generated the data sets by using the Markov chain Monte Carlo method on the generative NNBMs.

\subsection{Square grid NNBM with random biases}

Let us consider the generative NNBM on a $6\times 6$ square grid graph whose parameters are 
\begin{align}
b_i \underset{\mrm{i.i.d.}}{\sim}  \mcal{U}(-0.4,0.4),\quad w_{ii}=1,\quad w_{ij} = 0.8, 
\label{eq:SqNNBM}
\end{align}
where $\mcal{U}(\alpha,\beta)$ is the unique distribution from $\alpha$ to $\beta$.
Data set $\mcal{D}$ was generated from this generative NNBM, 
and used to train the learning NNBM.
Table \ref{tab:MAE-1} shows the MAEs obtained using the approximative methods.
\begin{table}[htbp]
\begin{center}
\caption{MAEs between true parameters and learning parameters in equation (\ref{eq:MAEs}). 
The complete data set $\mcal{D}$ are generated from the square grid NNBM in equation (\ref{eq:SqNNBM}).
Each result is the average over 50 trials.}
\label{tab:MAE-1}
\begin{tabular}{|c||c|c|c|} \hline
             &Downs~\cite{Downs2001}&SusP &I-SusP  \\ \hline
$e_{\bm{b}}$ &0.515      &0.297    &\textbf{0.153}   \\ \hline
$e_{\bm{w}}$ &0.376      &0.095    &\textbf{0.060}   \\ \hline
\end{tabular}
\end{center}
\end{table}
``SusP'' are the results obtained by the normal susceptibility propagation presented in section \ref{subsec:LRR&SusP}, 
and ``I-SusP'' are the results obtained by the improved susceptibility propagation presented in section \ref{subsec:I-SusP}.
``Downs`` are the results obtained by the mean-field learning algorithm proposed by Downs~\cite{Downs2001}.
We can see that the proposed improved method (I-SusP) outperformed the other methods.

\subsection{Neural network model for orientation tuning}

Let us consider a neural network model for orientation tuning in the primary visual cortex, 
which can be represented by a cooperative NNBM distribution~\cite{Ben-Yishai1995, Downs&Mackay&Lee2000, Downs2001}. 
The energy function of the neural network model is a translationally invariant function of the angles of maximal response of the $n$ neurons,
and can be mapped directly onto the energy of an NNBM whose parameters are 
\begin{align}
b_i = \beta,\quad w_{ij} = \beta\Big( \delta_{i,j} + \frac{1}{n} - \frac{\varepsilon}{n}\cos \big( 2\pi n^{-1}|i-j|\big)\Big),  
\label{eq:NN_NNBM}
\end{align}
where $\beta = 10$ and $\varepsilon = 2$.
Data set $\mcal{D}$ was generated from this generative NNBM, 
and used to train the learning NNBM.
Table \ref{tab:MAE-2} shows the MAEs obtained using the approximative methods.
\begin{table}[htbp]
\begin{center}
\caption{MAEs between true parameters and learning parameters in equation (\ref{eq:MAEs}). 
The complete data set $\mcal{D}$ was generated from the square grid NNBM in equation (\ref{eq:NN_NNBM}).
Each result is the average over 50 trials.}
\label{tab:MAE-2}
\begin{tabular}{|c||c|c|c|} \hline
             &Downs~\cite{Downs2001}&SusP &I-SusP  \\ \hline
$e_{\bm{b}}$ &10.772                 &1.644    &\textbf{0.480}   \\ \hline
$e_{\bm{w}}$ &0.700                  &0.151    &\textbf{0.107}   \\ \hline
\end{tabular}
\end{center}
\end{table}
We can see that the proposed improved method (I-SusP) again outperformed the other methods.

\section{Conclusion}\label{sec:conclusion}

In this paper, an effective inference method for NNBMs was proposed. 
This inference method was constructed by using the TAP equation and the diagonal consistency method, which was recently proposed, and 
performed better than the normal susceptibility propagation in the numerical experiments of statistical machine learning on NNBMs described in section \ref{sec:numerical}. 
Moreover, the order of the computational cost of the proposed inference method is the same as that of the normal susceptibility propagation, 
which is $O(n |E|)$.

The proposed method was developed based on the second-order perturbative approximation of the free energy in equation (\ref{eq:Gibbs_NNBM_2nd}). 
Since the Plefka expansion allows us to derive higher-order approximations of the free energy, 
we should be able to formulate improved susceptibility propagations base on the higher-order approximations in a similar way to the presented method. 
It should be addressed in future studies.
In reference \cite{Yasuda2007}, a TAP equation for more general continuous MRFs, where random variables are bounded within finite values, was proposed. 
We can formulate an effective inference method for continuous MRFs by combining the TAP equation with the diagonal consistency method 
in a way similar to that described in this paper.
This topic should be also considered in future studies.

\appendices

\section{brief derivation of variational free energy}\label{app:derivation-VFE}

To derive the variational free energy in equation (\ref{eq:GibbsFreeEnergy_NNBM}), let us consider the following functional with a trial probability density function $q(\bm{x})$:
\begin{align*}
f[q]:=\int_{0}^{\infty}E(\bm{x};\bm{b},\bm{w})q(\bm{x})\diff\bm{x}+\int_{0}^{\infty}q(\bm{x})\ln q(\bm{x})\diff\bm{x}, 
\end{align*}
and consider the minimization of this functional under constraints: 
\begin{align*}
m_i =\int_{0}^{\infty}x_i q(\bm{x})\diff\bm{x},\quad v_i =\int_{0}^{\infty}x_i^2 q(\bm{x})\diff\bm{x}.
\end{align*}
By using the Lagrange multipliers, this minimization yields equation (\ref{eq:GibbsFreeEnergy_NNBM}), i.e.,
\begin{align*}
\mcal{F}(\bm{m},\bm{v};\bm{b},\bm{w})=\min_{q}\big\{ f[q] \mid \mrm{constraints}\big\}.
\end{align*}
This variational free energy is referred to as the Gibbs free energy in statistical physics, 
and the minimum of the variational free energy coincides with $-\ln Z(\bm{b},\bm{w})$~\cite{Yasuda&Tanaka2012}.

\section*{Acknowledgment}
This work was partly supported by Grants-In-Aid (No. 24700220) 
for Scientific Research from the Ministry of Education, Culture, Sports, Science and Technology, Japan.

\end{document}